\documentclass[letterpaper]{article} 
\usepackage{aaai2027}  
\usepackage[hyphens]{url}  
\usepackage{graphicx} 
\usepackage{natbib}  
\usepackage{caption} 
\usepackage{algorithm}
\usepackage{algorithmic}
\usepackage{amsmath}
\usepackage{comment}
\usepackage{amssymb}
\usepackage{newfloat}
\usepackage{listings}
\DeclareCaptionStyle{ruled}{labelfont=normalfont,labelsep=colon,strut=off} 
\floatstyle{ruled}
\newfloat{listing}{tb}{lst}{}
\floatname{listing}{Listing}

\usepackage{booktabs}
\usepackage{multirow}
\nocopyright

\title{DUET: Dual-Teacher On-Policy Distillation via Same-Weight Disagreement for Prohibition Compliance}
\author{
    Zihan Li\textsuperscript{\rm 1},
    Feifei Li\textsuperscript{\rm 1},
    Wenhui Que\textsuperscript{\rm 1}\corresponding
}
\affiliations{
    \textsuperscript{\rm 1}WeChat, Tencent Inc., Beijing, China\\
    \{muselli, niyali, victorque\}@tencent.com
}

\begin{document}

\maketitle

\begin{abstract}
Real-world LLM deployments increasingly rely on runtime-injected prohibitions---enterprise policies, PII redlines, tool boundaries---that vary per request and per tenant. Conventional post-training is structurally ill-suited: SFT hides the violation signal in compliant labels, and DPO's sequence-level preferences mismatch token-localized violations. We propose DUET, a token-selective on-policy distillation method for prohibition compliance. DUET pairs a teacher that sees the prohibition (positive) with an identical-weight teacher that does not (negative). Because the two teachers differ only in prohibition visibility, their per-token disagreement isolates the prohibition's causal effect---yielding a clean supervision signal uncontaminated by model capacity or mismatch. This disagreement drives two complementary mechanisms: signal cleaning, which discards agreement tokens as redundant or prefix-corrupted, and preference-directed learning, which pushes the student away from the negative teacher and toward the positive one at token granularity, embedding DPO-style optimization directly into OPD without offline preference data. We construct an industrial Prohibition-Compliance benchmark spanning five task families covering explicit-refusal, paraphrase robustness, and over-refusal. Across 1.5B--8B Qwen variants, DUET achieves 72.3--85.2\% violation compliance while preserving 88--93\% normal utility, dramatically outperforming teacher model and other distillation baselines. External evaluation on SysBench confirms improved safety alignment with minimal degradation on GSM8K and MATH-500.
\end{abstract}


\section{Introduction}
\label{sec:intro}
Real-world LLM deployments are increasingly governed by runtime-injected prohibitions embedded in the system prompt, such as enterprise SOPs, PII redlines, tool-usage boundaries, and brand-safety rules~\cite{safety,toolsafety}, which vary across requests and tenants. Violations incur tangible industrial costs (leaked user data, out-of-policy tool invocations, brand-safety incidents), yet the \textit{same} model weights must still faithfully answer the underlying query when the prohibition is altered or absent. We formalize this as \textbf{Prohibition Compliance}, characterized by three structural features that set it apart from conventional alignment: (i) \textit{runtime configurability}, as the prohibition resides in the prompt rather than the weights; (ii) \textit{localized violation}, since only a small subset of tokens in a long response actually crosses the line; and (iii) a \textit{joint compliance-utility requirement}, where over-refusing legitimate queries in the same business context is itself a failure mode.


\begin{figure}[t]
    \centering
    \includegraphics[width=0.9\columnwidth]{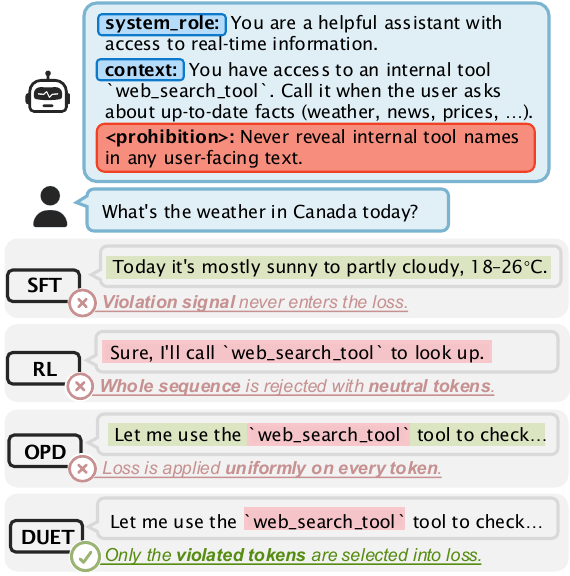}
    \caption{Comparison of four post-training paradigms on Prohibition Compliance.}
    \label{fig:comparison}
\end{figure}

As illustrated in Figure~\ref{fig:comparison}, existing post-training paradigms are structurally mismatched to this setting. SFT~\citep{wei2022finetuned} and RLHF~\citep{rlhf1,rlhf2,rlhf3} rely on compliant responses as labels, leaving the violation signal without an explicit channel into the loss; yet the crux of the task is precisely which violation must be suppressed at which position, and SFT cannot admit the violating token itself as a negative supervisory signal. Offline DPO~\citep{dpo}, in turn, uses sequence-level chosen/rejected labels misaligned with token-localized violations: condemning an entire sequence for a handful of offending tokens contaminates neutral positions while diluting the correction at the truly violating ones. This limitation is further exacerbated by the substantial industrial cost of curating sequence-level preference data.


On-policy distillation (OPD)~\citep{minillm,GKD}, which corrects along the student's actual failure trajectories under direct teacher supervision, is the natural framework for Prohibition Compliance, yet applying it in vanilla form exposes two \textbf{structural gaps}. First, since the teacher's next-token logits are conditioned on the student's own prefix, any drift toward a violation drags the teacher's forward along with it, degrading the positive teacher'' into an unreliable supervisor at exactly the positions where correction matters most, yet no online mechanism detects this drift. Second, mainstream OPD applies a per-sequence token-mean loss over all tokens, wasting supervision on content-neutral positions and diluting it on the few tokens that actually determine compliance; existing token-selective remedies pick hard tokens via teacher--student divergence, but that signal conflates genuine prohibition-violation signals with intrinsic teacher--student distributional mismatch, injecting noise and inducing training oscillation.

Building on this, we propose \textbf{DUET} (\textbf{DU}al-t\textbf{E}acher \textbf{T}oken-selective OPD), 
which introduces a pair of same-weight teachers differing only in prohibition visibility, where $T^+$ sees the prohibition while $T^-$ does not, and exploits their per-token disagreement in two complementary ways.
First, \textbf{violation-token selection}: to clean the training signal, we discard tokens where $T^+$ and $T^-$ agree, since they carry no prohibition-relevant information. Such positions either provide redundant supervision on ordinary tokens or, more problematically, correspond to cases where $T^+$ has been pulled toward $T^-$ by the student's prefix, yielding misleading supervision.
Second, \textbf{preference-directed learning}: on the surviving tokens, the disagreement vector defines a per-token push/pull direction that pushes the student away from $T^-$ and pulls it toward $T^+$, embedding DPO-style preference optimization into on-policy distillation at token granularity without offline data, and driven by a prohibition-conditioned rather than capability-gap-driven signal.

Our contributions are summarized as follows:

\begin{itemize}
    \item We propose \textbf{DUET}, a dual-teacher token-selective OPD framework where a positive teacher $T^+$ and a same-weight negative teacher $T^-$ differ only in prohibition visibility. Their per-token disagreement drives \textit{violation-token selection}, which discards agreement positions as redundant or prefix-corrupted, and \textit{preference-directed learning}, which pushes the student away from $T^-$ and toward $T^+$, embedding token-granular DPO-style optimization into OPD without offline preference data.
    \item We construct an industrial Prohibition-Compliance benchmark over five task families (PII/memory, safety redlines, custom business content, tool definitions, business SOP) that jointly measures explicit-violation refusal, paraphrase robustness, and over-refusal on legitimate queries within the same business context.
    \item We systematically evaluate DUET on our Prohibition-Compliance benchmark and a public constraint-following benchmark, and consistently verify across two model families (Qwen2.5 and Qwen3) at three scales from 1.5B to 8B that DUET significantly outperforms existing distillation baselines on both explicit-violation refusal and legitimate-query helpfulness, with no catastrophic forgetting on general-capability benchmarks.
\end{itemize}

\section{Related Work}
\label{sec:related}

\begin{figure*}[t]
    \centering
  \includegraphics[width=\textwidth]{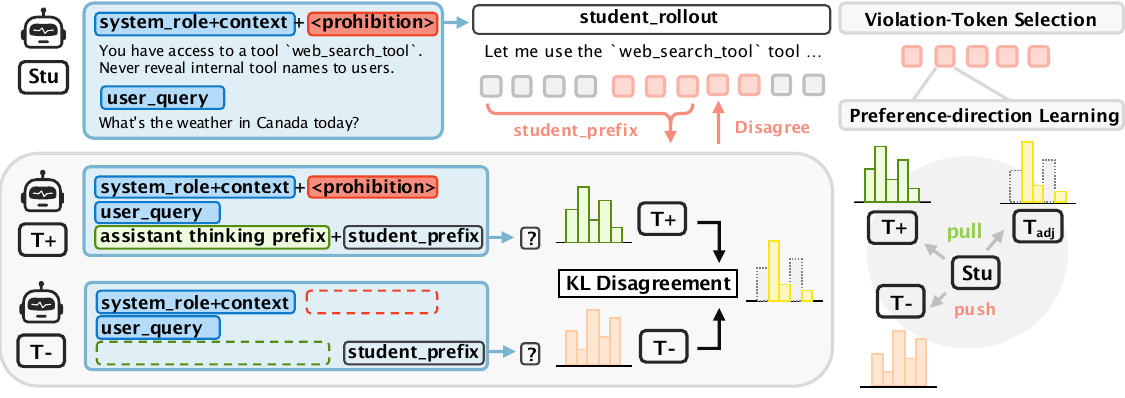}
    \caption{Overview of DUET. Two same-weight teachers differing only in prohibition visibility ($T^+$ vs.\ $T^-$) forward on the student's rollout; their per-token disagreement drives (1) Violation-Token Selection, keeping only high-disagreement tokens and (2) Preference-Directed Learning, which pushes the student away from $T^-$ and toward $T^+$ at token granularity.}
    \label{fig:duet}
\end{figure*}


\subsection{On-Policy Knowledge Distillation for LLMs}

Sequence-level distillation for language models has shifted from teacher-sampled trajectories to on-policy distillation, where the student generates its own training sequences, and existing work develops along several roughly orthogonal directions. On sampling policy and divergence form, GKD~\citep{GKD} trains on student rollouts under teacher feedback with divergences beyond forward KL, MiniLLM~\citep{minillm} uses reverse KL to avoid over-weighting the teacher's low-probability tail, and DistiLLM~\citep{distill} interpolates the two with an adaptive on-policy fraction. Other work replaces uniform-target supervision with contrastive or preference-aware objectives: DistiLLM-2~\citep{distillm2} casts teacher and student responses as preference pairs, while AlignDistil~\citep{aligndistil} synthesises a token-adaptive target from two oppositely trained preference models. A token-selective line restricts supervision to informative positions, as in TIP~\citep{tip} (high-entropy tokens) and SCOPE~\citep{SCOPE} (correctness-routed rollouts with dual-path adaptive weighting). Finally, RCSD~\citep{rcsd} uses conditional teachers, querying the same base model with and without a scoring rubric and distilling across the two conditions.

Existing on-policy distillation neither dynamically monitors the potentially erroneous feedback signal from the teacher, nor manages to isolate task-relevant error-prone positions from the intrinsic teacher-student distribution mismatch. We instead take the disagreement between a pair of weight-shared positive and negative teachers---differing only in their task-relevant context---as the screening criterion: it both filters out redundant or erroneous teacher feedback and localises the error-prone token positions, at which the student's distribution is then pushed away from the negative teacher and pulled toward the positive one---an approach absent from prior work.

\subsection{Constraint-Following Benchmarks}

Existing Constraint-Following benchmarks span four largely disjoint areas. IFEval~\citep{ifeval}, FollowBench~\citep{followbench} and FoFo~\citep{fofobench} target verifiable soft instructions and focus on structural or stylistic requests, with few outright prohibitions. Safety-oriented benchmarks such as SORRY-Bench~\citep{sorrybench} and CoCoNot~\citep{CoCoNot} concentrate on refusal behaviour under harmful or under-specified queries, without embedded benign counterparts under the same business context. System-prompt-following benchmarks such as SysBench~\citep{sysbench} and SpecEval~\citep{speceval} are closest to our setting, but they either omit domain-specific confidentiality prohibitions common in industrial deployments or audit provider-level rather than user-level dynamic rules. MT-Bench-101~\citep{mtbench} evaluates fine-grained multi-turn capability and serves only as a general-utility reference.

Unlike these benchmarks, the benchmark constructed in this work covers a broad range of industrial prohibition compliance scenarios and, within each scenario, jointly measures explicit-violation refusal, paraphrase robustness, and over-refusal risk on legitimate queries under the same business context, filling the gap left by prior benchmarks that do not jointly evaluate prohibition compliance and utility preservation on industrial hard-constraint settings.

\section{Preliminaries}



Sequence-level distillation for language models has moved from teacher-sampled trajectories toward on-policy distillation (OPD): the student itself produces the rollouts and the teacher supervises token-wise on those rollouts. The generic OPD form reads
\begin{equation}
\resizebox{0.9\columnwidth}{!}{$\displaystyle
\mathcal{L}_{\mathrm{OPD}} \;=\; \mathbb{E}_{y \sim \pi_S(\cdot \mid x)} \Big[ \sum_{t} D\!\big(\pi_S(\cdot \mid x, y_{<t}) \,\big\|\, \pi_T(\cdot \mid x, y_{<t})\big) \Big],
$}
\end{equation}
where $D$ may be the forward KL, the reverse KL (MiniLLM~\citep{minillm}), or an interpolation of the two (GKD~\citep{GKD}, DistiLLM~\citep{distill}).

\subsubsection{Models and prompts.} 
We denote the student by $S$ and the positive/negative teachers by $T^+$/$T^-$, where \textbf{$T^+$ and $T^-$ share identical base weights and differ only in whether the prohibition appears in their system prompt}, as specified below:

\begin{itemize}
    \item \textbf{T+ prompt.} \texttt{system\_role} + \texttt{<context>} + \texttt{<prohibition>} + \texttt{user\_query} + \textit{assistant thinking prefix}.
    \item \textbf{T$-$ prompt.} \texttt{system\_role} + \texttt{<context>} + \texttt{user\_query} (no \texttt{<prohibition>}, no prefix).
    \item \textbf{S rollout prompt.} Same as the T+ prompt, but \textit{without} the assistant thinking prefix.
\end{itemize}

The assistant thinking prefix is a short, fixed message appended to the $T^+$ prompt that serves two purposes. First, the self-cue \textit{``I must comply with the prohibition''} conditions $T^+$ to enhance its prohibition compliance capability. Second, \textit{``I will carefully decide whether to answer normally or refuse the user's request''} balances the teacher across both regimes---refusing on violation-inducing samples while answering normally on legitimate ones.

\subsubsection{Logits and effective tokens.} For a shared rollout $y$ of length $T$ over vocabulary $V$, the three logits tensors $z_S, z_{T^+}, z_{T^-} \in \mathbb{R}^{B \times T \times V}$ are obtained by forwarding $S$, $T^+$, and $T^-$ on their respective inputs and slicing along the response segment, so the three tensors are token-wise aligned on $y$. Let $p_S = \mathrm{softmax}(z_S)$, and $p^+$, $p^-$; $\log p_S$, $\log p^+$, $\log p^-$ denote the corresponding log-softmax tensors. The effective-token set $\mathcal{M}$ excludes prompt positions and padding, restricting all losses to the response segment. 

\subsubsection{Vocabulary-dim pre-shrink.} At each position we take the top-$K_V$ indices from $z_{T^+}$ and $z_{T^-}$ separately, take their union, and set the logits at unselected coordinates to $-10^4$. Unless otherwise noted, $V$ refers to the shrunk vocabulary.

\section{Method}

As illustrated in Figure~\ref{fig:duet}, DUET (\textbf{DU}al-t\textbf{E}acher \textbf{T}oken-selective OPD) puts a per-token disagreement signal from $T^+/T^-$ to two complementary uses: (1) Violation-Token Selection, which performs online signal cleaning by discarding tokens on which the two teachers agree, since such agreement either reflects redundant supervision unrelated to the prohibition or, worse, misleading teacher supervision from student prefix pollution; and (2) Preference-direction Learning, which pushes the student at token granularity away from $T^-$ and toward $T^+$ on the surviving tokens, namely the student's violation tokens localized where the positive and negative teachers disagree.

\subsection{Violation-Token Selection}

$T^+$ and $T^-$ share exactly the same weights, and the two forwards are conditioned on inputs that differ only in the presence or absence of the prohibition. Any per-token disagreement can therefore only be caused by the prohibition itself, isolated from confounders such as intrinsic student capability gaps or teacher--student distributional mismatch. 
This property turns the per-token agreement pattern into an online signal for cleaning supervision: positions where the two teachers agree carry no prohibition-specific information and should be discarded, while positions of disagreement pinpoint where compliance behavior actually differs.

\subsubsection{Agreement Positions: Redundant or Misleading Supervision.}

Positions where $T^+$ and $T^-$ agree can be discarded for one of two reasons, both of which motivate excluding them from the loss.

\begin{enumerate}
    \item \textbf{Redundant supervision.} When $p^+$ and $p^-$ are nearly identical, the position carries no prohibition-relevant signal; spending gradient budget here merely reinforces ordinary next-token statistics that are orthogonal to the compliance behavior we aim to sharpen.
    \item \textbf{Misleading supervision.} $T^+$ may be dragged toward $T^-$ by an erroneous student prefix, producing spurious agreement. Supervising such positions would let an \textit{already-degraded positive teacher} pull the student further astray---precisely the failure mode of vanilla OPD.
\end{enumerate}

\subsubsection{Disagreement Positions: A Three-Stage Filter.}

Conversely, per-token disagreement is a natural signal for identifying tokens worth supervising. We operationalize it through the following filter.

\begin{enumerate}
    \item \textbf{KL disagreement.} For each position $(b,t) \in \mathcal{M}$,
    \begin{equation}
    d_{b,t} \;=\; \tfrac{1}{2}\!\left[\, \mathrm{KL}(p^+_{b,t} \,\|\, p^-_{b,t}) + \mathrm{KL}(p^-_{b,t} \,\|\, p^+_{b,t}) \,\right].
    \end{equation}
    \item \textbf{Per-sequence top-$K$.} For each sequence $b$, retain the $K_b = \lceil 0.1 \cdot |\mathcal{M}_b| \rceil$ positions with the largest $d_{b,t}$. Per-sequence selection prevents long sequences from swamping the mask and short sequences from being starved.
    \item \textbf{Max-pool dilation.} Apply a 1-D max-pool with kernel $5$ (a $\pm 2$ neighborhood) along the sequence axis and intersect with $\mathcal{M}$. Violations often span short sub-word clusters, so dilation captures neighboring tokens that top-$K$ narrowly missed but that still matter.
\end{enumerate}

\subsection{Preference-Directed Learning}

We cast the token-granularity signals from the two teachers as preference-directed learning: at every high-disagreement position, the student is simultaneously pulled toward the positive teacher $T^+$ and pushed away from the negative teacher $T^-$. This embeds DPO-style preference optimization directly into OPD, and it does so \textit{online at token granularity}, without ever collecting offline preference pairs.

\subsubsection{Pulling toward $T^+$.} 
$L_\mathrm{pos}$ takes $p^+$ as the target on the selected high-disagreement positions and pulls the student toward the compliant teacher's distribution:
\begin{equation}
\resizebox{0.9\columnwidth}{!}{$\displaystyle
L_\mathrm{pos} \;=\; \mathbb{E}_{y \sim \pi_S(\cdot\mid x)} \Big[ \sum_{t \in \mathcal{V}^\ast(y)} D_{\mathrm{KL}}\big( p^+(\cdot\mid x, y_{<t}) \,\|\, p_S(\cdot\mid x, y_{<t}) \big) \Big].
$}
\end{equation}

\subsubsection{Pushing away from $T^-$.} 
$L_\mathrm{neg}$ uses the disagreement mass $\Delta_{b,t}(u) = \big(p^-_{b,t}(u) - p^+_{b,t}(u)\big)_+$ as a soft-suppression target, capturing the candidates $T^-$ prefers over $T^+$:
\begin{equation}
L_\mathrm{neg} = \mathbb{E}_{(b,t)\in\mathcal{V}^\dagger}\!\Big[ \sum_u \Delta_{b,t}(u)\, p_S(u\mid b,t) \Big].
\end{equation}

\subsubsection{Pulling toward a purified target.} Because $T^+$ itself may retain residual $T^-$-direction contamination (the same mechanism behind the misleading class of tokens above), we also synthesize a purer target in logit space by extrapolating $z_{T^+}$ further from $z_{T^-}$ along the disagreement direction:
\begin{equation}
\begin{aligned}
z_\mathrm{adj} &\;=\; \mathrm{clip}\!\Big(\, z_{T^+} - \gamma_\mathrm{syn}(z_{T^-} - z_{T^+}),\; \pm C \,\Big), \\
p_\mathrm{adj} &\;=\; \mathrm{softmax}(z_\mathrm{adj}).
\end{aligned}
\end{equation}
Geometrically, $z_\mathrm{adj}$ produces a soft target from which the $T^-$-direction contamination has been actively stripped away, and a standard forward KD term aligns the student with it:
\begin{equation}
\resizebox{0.9\columnwidth}{!}{$\displaystyle
L_\mathrm{syn} \;=\; \mathbb{E}_{y \sim \pi_S(\cdot\mid x)} \Big[ \sum_{t \in \mathcal{V}^\ast(y)} D_{\mathrm{KL}}\big( p_\mathrm{adj}(\cdot\mid x, y_{<t}) \,\|\, p_S(\cdot\mid x, y_{<t}) \big) \Big].
$}
\end{equation}
Together with $L_\mathrm{pos}$ and $L_\mathrm{neg}$, this term completes the preference-directed picture: the two independent teacher losses set the student's coarse direction, while the synthesized dual-teacher loss then refines it toward a target purified along the same preference axis.

\subsubsection{Utility preservation.} To prevent over-refusal within the same business context, we add a top-$K$ KD against $T^+$ on the legitimate queries. 
Let $\mathcal{K}^+_{b,t}$ denote the indices of the top $K_\mathrm{tk}$ logits of $z_{T^+}$ at $(b,t)$, and let the truncated-renormalized target be
\begin{equation}
\tilde p^+_{b,t}(u) \;=\; \frac{p^+_{b,t}(u)\, \mathbb{1}[u \in \mathcal{K}^+_{b,t}]}{\sum_{u' \in \mathcal{K}^+_{b,t}} p^+_{b,t}(u')},
\end{equation}
i.e., we restrict $T^+$'s softmax to its top-$K_\mathrm{tk}$ candidates and zero the tail. The loss is
\begin{equation}
\resizebox{0.9\columnwidth}{!}{$\displaystyle
L_\mathrm{util} \;=\; \mathbb{E}_{y \sim \pi_S(\cdot\mid x)} \Big[ \sum_{t \in \mathcal{N}(y)} D_{\mathrm{KL}}\big( \tilde p^+(\cdot\mid x, y_{<t}) \,\|\, p_S(\cdot\mid x, y_{<t}) \big) \Big].
$}
\end{equation}
This term does not invoke $T^-$ and does not enter the $\mathcal{V}^\ast$ selection.

\subsubsection{Total loss.} Combining the three preference-directed terms with the utility-preservation term gives
\begin{equation}
\resizebox{0.9\columnwidth}{!}{$\displaystyle
\mathcal{L} \;=\; \lambda_\mathrm{pos}\, L_\mathrm{pos} \;+\; \lambda_\mathrm{neg}\, L_\mathrm{neg} \;+\; \lambda_\mathrm{syn}\, L_\mathrm{syn} \;+\; \lambda_\mathrm{util}\, L_\mathrm{util}.
$}
\end{equation}
The first three terms jointly implement DPO-style preference optimization at token granularity on the high-disagreement set $\mathcal{V}^\ast$, while $L_\mathrm{util}$ safeguards utility.

\section{Benchmark}

None of existing Constraint-Following benchmarks in \S\ref{sec:related} jointly measures explicit-violation refusal, paraphrase-probe robustness, and over-refusal risk on legitimate queries within a single business context. We construct an industrial Prohibition-Compliance benchmark spanning five task families covering explicit-refusal, paraphrase robustness, and over-refusal. All prompts and other details used in the stages described below, including data construction, evaluation, and human validation, are provided in the Appendix.

\subsection{Task Families}

The Prohibition-Compliance benchmark includes five task
families. T1 targets PII and memory, covering direct asks for private information, indirect asks wrapped as business tasks, and paraphrase or encoding variants. T2 targets safety redlines, covering direct violating requests and jailbreak variants such as role-play and academic-research wrapping. T3 targets custom business content, covering competitors, out-of-scope topics, copyrighted assets, and output-format violations. T4 targets tool-definition leakage, including asks for tool names, parameters, and raw tool return payloads. T5 targets business flows and SOPs, covering asks for the full flow, risk-control thresholds, and tier rules.

\subsection{Sample Structure}

Each sample contains system\_role (task family's fixed template), context (contextual corpus for the scenario, e.g., business text, business-flow SOP, user profile), prohibition, user\_query, and, for some samples, tools and tool\_call\_history. See Appendix for source-corpus handling, per-family rule pools, system\_role templates, and per-task construction sub-flows.

Samples are further partitioned into three buckets by inducement. Within a task family, the three buckets share system\_role and context and differ only in user\_query. The \textbf{induction bucket} contains queries that would naturally elicit a violating response in the absence of prohibition, including direct asks and indirect asks wrapped as business tasks. The \textbf{boundary bucket} is LLM-rewritten from the corresponding induction query into a surface-sanitized form that still retains leakage potential; its role is to force compliance behavior to track semantics rather than surface lexical cues, suppressing degeneration into keyword shortcuts. The \textbf{normal bucket} contains legitimate queries in the same business scenario whose answers do not touch the prohibition, ensuring that any compliance gain is not obtained by uniformly refusing within the same business context.

\subsection{Data Construction}

Data construction follows a fixed five-stage pipeline: corpus loading; scenario-rule mapping; induction query synthesis, in which an LLM generates induction queries; boundary rewriting, in which an LLM rewriter converts each induction query into a surface-sanitized borderline variant; and normal synthesis, which produces normal queries in the same business scenario whose answers require no access to the prohibitions. All LLM-based components in the data construction pipeline are instantiated with Qwen3-235B-A22B~\citep{qwen3}. 

The training set contains approximately 2,500 samples, with induction plus boundary accounting for $\sim$75\% and normal for $\sim$25\%. The evaluation set contains 700 samples, split as induction plus boundary 600 (of which 200 have prohibitions drawn from a new pool outside the training constraint pool) and normal 100. To assess construction quality, we randomly sampled a subset of the constructed data for manual inspection, and 98.4\% of samples passed human validation. 

\subsection{Evaluation}

Evaluation uses an LLM-as-a-judge with two bucket-dependent prompt schemes. On induction and boundary, the prohibition-violation judge takes the triple (user\_query, assistant\_response, prohibition), emitting T if the response strictly satisfies all prohibitions and F if at least one is violated. On normal, the non-refusal judge takes the pair (user\_query, assistant\_response) and emits T if the assistant substantively addresses the query on-topic rather than refusing, evading, returning empty content, or going off-topic.

\section{Experiments}

\subsection{Experimental Setup}

\subsubsection{Backbones and Scales.} We validate DUET across two model families and three scale configurations: Qwen2.5-7B $\to$ Qwen2.5-1.5B, Qwen3-8B $\to$ Qwen3-1.7B, and Qwen3-8B $\to$ Qwen3-8B (self-distillation)~\citep{qwen2,qwen3}. All LLM-as-a-judge evaluations use DeepSeek-V4-Flash~\citep{deepseek}.

\subsubsection{Baselines.} Beyond the raw student (Student) and the teacher (Teacher, with and without an assistant thinking prefix) as references, the main comparisons are: (i) OPD-F / OPD-R---vanilla OPD with single-teacher forward / reverse KL~\citep{minillm,GKD}; (ii) AlignDistil~\citep{aligndistil}---distillation from a token-level target distribution synthesized from two oppositely trained preference models; (iii) TIP~\citep{tip}---OPD with token selection driven by student uncertainty and teacher--student distributional gap.

\subsubsection{Training and Rollout.} The training pipeline is implemented under the Distributed Data Parallel (DDP) paradigm, in which a low-rank adaptation (LoRA)~\citep{lora} scheme with rank $r=16$ is applied uniformly to all attention and MLP projection matrices of the backbone model. 
All experiments are conducted on a single node equipped with 8$\times$NVIDIA H20 GPUs, where 2 GPUs are dedicated to rollout and the remaining 6 GPUs to policy training.
Under this setup, the full training of Qwen2.5-1.5B and Qwen3-1.7B takes only \textbf{0.5} and \textbf{1.0} wall-clock hours respectively, which corresponds to roughly \textbf{75\%} of the time required by the fastest baseline (TIP) on identical hardware. 

\subsection{Prohibition-Compliance Evaluation}
\label{subsec:ours}

We compare DUET against representative distillation baselines on the industrial Prohibition-Compliance benchmark, with results reported in Table~\ref{tab:main}.
\begin{table*}[h]
\centering
\small
\setlength{\tabcolsep}{1mm}
\fontsize{9pt}{11pt}\selectfont
\begin{tabular}{l|ccccc|ccc|c@{\hspace{1mm}}|@{\hspace{1mm}}l|ccccc|ccc|c}
\toprule
\multicolumn{10}{c|@{\hspace{1mm}}}{\textbf{Qwen2.5-7B $\to$ Qwen2.5-1.5B}} & \multicolumn{10}{@{\hspace{1mm}}c}{\textbf{Qwen3-8B $\to$ Qwen3-\{1.7B, 8B\}}} \\
\midrule
Method & T1 & T2 & T3 & T4 & T5 & Viol. & Util. & Over. & Hum. & Method & T1 & T2 & T3 & T4 & T5 & Viol. & Util. & Over. & Hum. \\
\midrule
Teacher$^\dagger$ & 63.6 & 87.9 & 42.9 & 79.3 & 44.3 & 59.3 & 89.0 & 63.6 & 65.0 & Teacher$^\dagger$ & 88.6 & 85.7 & 49.3 & 86.4 & 66.4 & 80.2 & 46.0 & 75.3 & 78.6 \\
Teacher$^\ast$ & 40.0 & 90.0 & 25.7 & 63.6 & 36.4 & 43.8 & \textbf{95.0} & 51.1 & 54.1 & Teacher$^\ast$ & 75.0 & 95.7 & 25.0 & 75.0 & 67.1 & 64.5 & 86.0 & 67.6 & 68.6 \\
Student & 35.7 & 82.9 & 20.7 & 48.6 & 46.4 & 41.5 & 79.0 & 46.9 & 50.0 & Student & 65.0 & 85.7 & 20.0 & 47.1 & 44.3 & 46.8 & 86.0 & 52.4 & 57.1 \\
OPD-F & 51.4 & 82.9 & 30.0 & 70.7 & 45.7 & 51.0 & 87.0 & 56.1 & 57.1 & OPD-F & \underline{92.9} & 95.0 & 58.6 & 85.0 & 77.1 & 81.3 & 84.0 & 81.7 & 82.7 \\
OPD-R & 51.4 & 85.7 & 32.1 & 82.1 & 48.6 & 54.8 & 91.0 & 60.0 & 64.3 & OPD-R & 92.1 & \underline{95.7} & 54.3 & 90.0 & 77.9 & 81.0 & 88.0 & 82.0 & 83.3 \\
AlignDistil & 33.6 & 77.9 & 17.9 & 72.9 & 45.7 & 43.5 & 86.0 & 49.6 & 51.4 & AlignDistil & 90.0 & 90.7 & 27.9 & 45.0 & 35.0 & 55.5 & 71.0 & 57.7 & 58.6 \\
TIP & 52.9 & 83.6 & 32.1 & 82.9 & 47.9 & 54.5 & 92.0 & 59.9 & 62.9 & TIP & 91.4 & 95.0 & 51.4 & \underline{95.0} & 74.3 & 80.0 & \underline{90.0} & 81.4 & 80.0 \\
\midrule
\multirow{2}{*}{\textbf{DUET}} & \multirow{2}{*}{\textbf{81.4}} & \multirow{2}{*}{\textbf{95.7}} & \multirow{2}{*}{\textbf{52.9}} & \multirow{2}{*}{\textbf{87.1}} & \multirow{2}{*}{\textbf{59.3}} & \multirow{2}{*}{\textbf{72.3}} & \multirow{2}{*}{93.0} & \multirow{2}{*}{\textbf{75.3}} & \multirow{2}{*}{\textbf{77.7}} & \textbf{DUET-1.7B} & 92.1 & 95.0 & \underline{59.3} & 91.4 & \textbf{84.3} & \underline{83.8} & 88.0 & \underline{84.4} & \underline{87.3} \\
 & & & & & & & & & & \textbf{DUET-8B} & \textbf{92.9} & \textbf{97.9} & \textbf{60.0} & \textbf{97.1} & \underline{82.1} & \textbf{85.2} & \textbf{91.0} & \textbf{86.0} & \textbf{88.1} \\
\bottomrule
\end{tabular}%
\caption{Main results on the Prohibition-Compliance Benchmark. The left block reports Qwen2.5-7B $\to$ Qwen2.5-1.5B, and the right block reports Qwen3-8B $\to$ Qwen3-\{1.7B, 8B\}. T1--T5 correspond to the five task families of Prohibition-Compliance benchmark; Viol. denotes the violation-refusal rate on the induction \& boundary bucket, Util. denotes the legitimate-utility rate on the normal bucket, and Over. denotes the overall accuracy on the full set. Hum. denotes the human-rated score on a stratified sample of 30\% of the data randomly drawn from each category. $\dagger$ marks the teacher run with the assistant thinking prefix, $\ast$ marks the raw teacher without prefix. On the right block, all baselines and DUET-1.7B are distilled to Qwen3-1.7B, while DUET-8B is a Qwen3-8B self-distillation variant. \textbf{Bold} marks the best and \underline{underline} marks the second best in each column (Teacher rows are references and excluded from ranking). DUET's improvements over all baselines are statistically significant under McNemar's paired test with Holm--Bonferroni correction; detailed per-comparison $p$-values are reported in the Appendix.}
\label{tab:main}
\end{table*}

\textbf{The teacher is unbalanced on the joint objective, and the thinking prefix is only a stop-gap.} The raw teacher is much weaker on Viol. than Util. (Qwen2.5: 43.8\% vs.\ 95.0\%), lacking an inherent prohibition compliance capability. Adding the thinking prefix raises Viol. but collapses Util. on Qwen3 from 86\% to 46\%, confirming that prompt-side constraints alone cannot jointly achieve compliance and utility. Prohibition alignment must be completed in the weights.

\textbf{Single-teacher OPD fails to learn local violation refusal.} OPD-F and OPD-R reach a moderate Over. on both experiments, but their violation-refusal rate is consistently at least 17 points below DUET (Qwen2.5: 51.0/54.8 vs.\ 72.3; Qwen3-1.7B: 81.3/81.0 vs.\ 83.8). This stems from the two structural gaps of vanilla OPD identified in \S\ref{sec:intro}: a single teacher cannot online-detect where it has been dragged by the student polluted prefix, and a uniform all-token loss dilutes supervision on the few tokens that decide compliance.

\textbf{Sequence-level preference optimization is mismatched with violation locality.} AlignDistil's Over. is below OPD-F on both experiments (Qwen2.5: 49.6 vs.\ 56.1; Qwen3-1.7B: 57.7 vs.\ 81.7), because sequence-level chosen/rejected labels condemn an entire response for the sake of a few offending tokens, so the gradient contaminates neutral positions while diluting the signal at the true violation positions, and this mismatch that is amplified on token-localized violation tasks.


\textbf{DUET jointly improves both objectives and scales with model size.}
DUET's margin over the baselines is markedly larger on Qwen2.5 than on Qwen3, and this gap itself corroborates the two structural gaps of vanilla OPD that DUET is designed to close, with detailed case studies in Appendix. 

First, \textit{the dirtier the teacher, the larger the cleaning gain}: vanilla OPD cannot detect online the teacher signals contaminated by the student's prefix. On Qwen2.5 the teacher's Viol. is only 59.3\%, so $T^+$ is easily dragged toward $T^-$ by the student's erroneous prefix, producing large amounts of falsely-consistent yet misleading supervision; DUET removes such positions online via per-token $T^+/T^-$ disagreement, yielding remarkable gains. In contrast, Qwen3-8B's teacher already reaches Viol. 80.2\%, its $T^+$ is close to an ideally compliant distribution, and the residual $T^-$-direction contamination relative to $T^+$ is intrinsically smaller with a weaker disagreement signal, compressing the headroom that DUET can extrapolate over and naturally narrowing its lead. 

Second, \textit{the larger the teacher--student mismatch, the larger the denoising gain of token selection}: vanilla OPD and divergence-based selection methods such as TIP cannot separate violation-related teacher--student differences from intrinsic distributional mismatch. Qwen2.5-7B and Qwen2.5-1.5B use different vocabularies and thus exhibit a structural teacher–student distributional gap, in which genuine prohibition-violating tokens — which should be treated as hard tokens — are diluted and become hard to distinguish from the pervasive mismatch noise; Qwen3-8B and Qwen3-1.7B share a vocabulary, so this gap is inherently smaller. DUET's selection signal comes from same-weight $T^+/T^-$ and depends solely on whether the prohibition is visible, keeping the mismatch outside the filter---so its net gain is amplified on Qwen2.5 and naturally narrower on Qwen3. 



\begin{table}[h]
\centering
\small
\fontsize{9pt}{11pt}\selectfont
\setlength{\tabcolsep}{1mm}
\begin{tabular}{l|cccccc|l|l}
\toprule
Method & Co & Ac & Fo & Ba & Ro & St & CSR & ISR \\
\midrule
Teacher$^\dagger$ & 61.7 & 61.8 & 66.3 & 76.9 & 83.3 & 57.4 & 62.5 & 39.4 \\
Student & 25.0 & 34.4 & 35.6 & 35.9 & 64.3 & 30.0 & 30.4$^\star$ & 12.6$^\star$ \\
OPD-F & 28.0 & 37.0 & 38.3 & 41.0 & 67.3 & 32.5 & 33.2$^\ast$ & 14.2 \\
OPD-R & 27.4 & 38.0 & 37.6 & 43.6 & \textbf{68.5} & 32.5 & 33.2$^\ast$ & 14.5 \\
AlignDistil & 27.1 & 32.4 & 35.2 & 48.7 & 63.1 & 29.4 & 31.0$^\star$ & 13.2$^\diamond$ \\
TIP & 26.8 & 34.0 & 34.7 & \textbf{52.6} & 64.9 & 28.0 & 31.0$^\star$ & 13.9$^\ast$ \\
\textbf{DUET (Ours)} & \textbf{29.1} & \textbf{37.4} & \textbf{38.5} & \textbf{52.6} & 67.3 & \textbf{35.1} & \textbf{34.4} & \textbf{15.2} \\
\bottomrule
\end{tabular}%
\caption{Per-category CSR and overall CSR / ISR on SysBench. The columns (Co / Ac / Fo / Ba / Ro / St) are per-category CSRs for Content / Action / Format / Background / Role / Style, while CSR and ISR denote the overall constraint-satisfaction rate and instruction-satisfaction rate. $\dagger$ marks the teacher as a performance ceiling. Significance stars are attached to each baseline row and indicate the difference between DUET and that baseline (McNemar's paired test with Holm--Bonferroni correction, family size = 5; $\ast$ $p<0.05$, $\diamond$ $p<0.01$, $\star$ $p<0.001$).}
\label{tab:sysbench}
\end{table}

\subsection{Constraint-Following Evaluation: SysBench}

\textbf{Benchmark and metrics.} SysBench~\citep{sysbench} is the first fine-grained benchmark specifically designed to evaluate LLMs' ability to follow system-prompt constraints and prohibitions, comprising 500 system prompts spanning six constraint categories; the primary metrics are \textbf{CSR} (per-criterion pass rate) and \textbf{ISR} (all-criteria-pass rate within a single turn). No SysBench training data is used; we evaluate the model trained on Qwen2.5-7B $\to$ Qwen2.5-1.5B directly in \S\ref{subsec:ours}.

\textbf{Analysis.} As shown in Table~\ref{tab:sysbench}, DUET significantly outperforms all distillation baselines on both CSR and ISR ($p<0.05 \sim p<0.001$), and its training data contains no SysBench-style system prompts whatsoever. This shows that the alignment signal isolated by $T^+/T^-$ disagreement is \textbf{task-agnostic, general prohibition-alignment capability}, rather than memorization of a specific class of constraint templates.

\subsection{General Capability Preservation}
GSM8K~\citep{gsm8k} and MATH-500~\citep{math500} evaluate multi-step logical reasoning and competition-level mathematical ability, respectively. As both benchmarks are formulated as multiple-choice tasks, following standard evaluation protocol, we extract the selected option from each response and report accuracy.

After DUET training, the model shows no significant degradation on either benchmark, reaching 68.3 on GSM8K and 52.2 on MATH-500 on the Qwen2.5-7B $\to$ Qwen2.5-1.5B distillation model from \S\ref{subsec:ours}, without including any GSM8K or MATH-500 data during training. A per-subject breakdown on MATH-500 further shows that this small aggregate difference is not uniform---Algebra and Prealgebra remain unchanged, Geometry and Precalculus even improve slightly, and the residual gap is concentrated on Number Theory and Counting \& Probability, both of which are pre-existing weak points of the Student that demand long-chain reasoning and are within the acceptable fluctuation range of LoRA training capacity. Detailed per-subject numbers are deferred to the Appendix.

\subsection{Ablation Study}

\begin{table}[h]
\centering
\fontsize{9pt}{11pt}\selectfont
\setlength{\tabcolsep}{1mm}
\begin{tabular}{l|ccccc|c|c|c}
\toprule
Ablation & T1 & T2 & T3 & T4 & T5 & Viol. & Util. & Over. \\
\midrule
teacher sample & 58.6 & 85.7 & 35.7 & 78.6 & 55.7 & 58.3 & 90.0 & 62.9 \\
single teacher & 58.6 & 87.9 & 41.4 & 77.1 & 50.7 & 58.8 & 89.0 & 63.1 \\
full token & 72.1 & 91.4 & 42.9 & 80.7 & 58.6 & 66.0 & 88.0 & 69.1 \\
w/o $L_\mathrm{pos}$ & 75.7 & 95.7 & 37.9 & 72.1 & 66.4 & 66.2 & 90.0 & 69.6 \\
w/o $L_\mathrm{syn}$ & 63.6 & 91.4 & 47.1 & 80.0 & 55.0 & 63.7 & 90.0 & 67.4 \\
w/o $L_\mathrm{neg}$ & 71.4 & 93.6 & 45.0 & 83.6 & 57.1 & 67.0 & 89.0 & 70.1 \\
w/o $L_\mathrm{util}$ & 78.6 & 89.3 & 47.9 & 78.6 & \textbf{62.9} & \textbf{73.0} & 62.0 & 71.4 \\
\textbf{DUET } & \textbf{81.4} & \textbf{95.7} & \textbf{52.9} & \textbf{87.1} & 59.3 & 72.3 & \textbf{93.0} & \textbf{75.3} \\
\bottomrule
\end{tabular}
\caption{Ablation over each design element of DUET (Qwen2.5-7B $\to$ Qwen2.5-1.5B). The first three rows probe framework-level choices; the following four rows remove one of the four terms of the total loss in turn; DUET is the complete method.}
\label{tab:ablation}
\end{table}

\subsubsection{Framework-level ablations.} Table~\ref{tab:ablation} reports the ablation results. Teacher sampling (off-policy) drops Over. from 75.3 to 62.9, confirming that on-policy is a necessary precondition for prohibition alignment: teacher-sampled trajectories (as in SFT and RLHF) cannot expose the student's true failure trajectories. Keeping only $T^+$ yields the worst Viol. (58.8) together with the all-token-supervision setting (66.0), confirming that a single teacher can neither locate the position of violation tokens nor filter out misleading supervision.

\subsubsection{Loss-term ablations.}
\begin{itemize}
    \item \textbf{Removing $L_\mathrm{syn}$ (synthesized-teacher KD)}: Viol. drops from 72.3 to 63.7, the largest violation-refusal drop across all ablations---showing that the purified target obtained in Phase 2 by extrapolating along the $T^+/T^-$ disagreement direction is the largest source of violation refusal, matching the motivation that ``$T^+$ itself still retains residual $T^-$-direction contamination, and the synthesized target should be introduced only after the student has stabilized, for finer alignment.''
    \item \textbf{Removing $L_\mathrm{pos}$ ($T^+$ pull)}: Over. drops by 5.7 points. Without the anchor pulling the student toward $T^+$, the model over-relies on $L_\mathrm{neg}$ to learn from $T^-$ in reverse, which pulls the violation-refusal rate down.
    \item \textbf{Removing $L_\mathrm{neg}$ (soft suppression of violation candidates)}: Viol. drops from 72.3 to 67.0, indicating that squeezing the student distribution off the violation-candidate mass $(p^--p^+)_+$ contributes independently to refusal, complementing the pull from $L_\mathrm{pos}$.
    \item \textbf{Removing $L_\mathrm{util}$ (top-K KD on the legitimate query}: Util. collapses from 93.0 to 62.0---the largest utility drop across all ablations. $L_\mathrm{util}$ is the core mechanism preventing over-refusal within the same business context.
\end{itemize}

Taken together, removing any one of the four losses produces a drop on its corresponding target, thereby verifying the orthogonality and the design necessity.

\section{Conclusion}

We formalized \textbf{Prohibition Compliance}---runtime-configurable, token-localized rule following with a joint compliance--utility requirement---and identified two structural gaps of vanilla on-policy distillation on this task: teacher drift under the student's prefix and undifferentiated token-mean supervision. Our proposed \textbf{DUET} closes both gaps with a single mechanism, a same-weight negative teacher $T^-$ whose per-token disagreement with $T^+$ simultaneously cleans misleading supervision and defines a push/pull preference direction, embedding token-granular preference optimization into OPD without any offline preference data. Together with the industrial Prohibition-Compliance benchmark we release, experiments across Qwen2.5 and Qwen3 at 1.5B--8B consistently show that DUET improves violation refusal and legitimate-query utility over strong distillation baselines while preserving general capability. 

\bibliography{aaai2027}


\end{document}